\documentclass[letterpaper, 10 pt, conference]{ieeeconf}  

\IEEEoverridecommandlockouts                              

\usepackage{graphics} 
\usepackage{epsfig} 
\usepackage{mathptmx} 
\usepackage{times} 
\usepackage{amsmath} 
\usepackage{amssymb}  
\usepackage{afterpage}
\usepackage{newtxmath}

\usepackage{graphicx}
\usepackage{upgreek}
\usepackage{bbm}
\usepackage{dsfont}
\usepackage{caption}
\usepackage{subcaption}
\usepackage{color}

\usepackage{booktabs}
\usepackage{makecell}
\usepackage{titlesec}
\usepackage{wrapfig}
\usepackage{array}
\usepackage{multirow}
\usepackage{colortbl}
\usepackage{tabularx}
\usepackage{cite}

\usepackage{xcolor}
\usepackage{pifont}
\usepackage{hyperref}
\usepackage{bm}

\definecolor{gold}{RGB}{255,215,0}
\definecolor{silver}{RGB}{200, 200, 255}
\newcommand{\best}[1]{\cellcolor{gold!40}\textbf{#1}}
\newcommand{\second}[1]{\cellcolor{silver!40}#1}
\definecolor{lightblue}{RGB}{80,150,255}

\begin{document}

\thispagestyle{empty}

\makeatletter
\let\@oldmaketitle\@maketitle
\renewcommand{\@maketitle}{\@oldmaketitle 
  \centering
  \setcounter{figure}{0}%
  \includegraphics[width=1.0\linewidth]{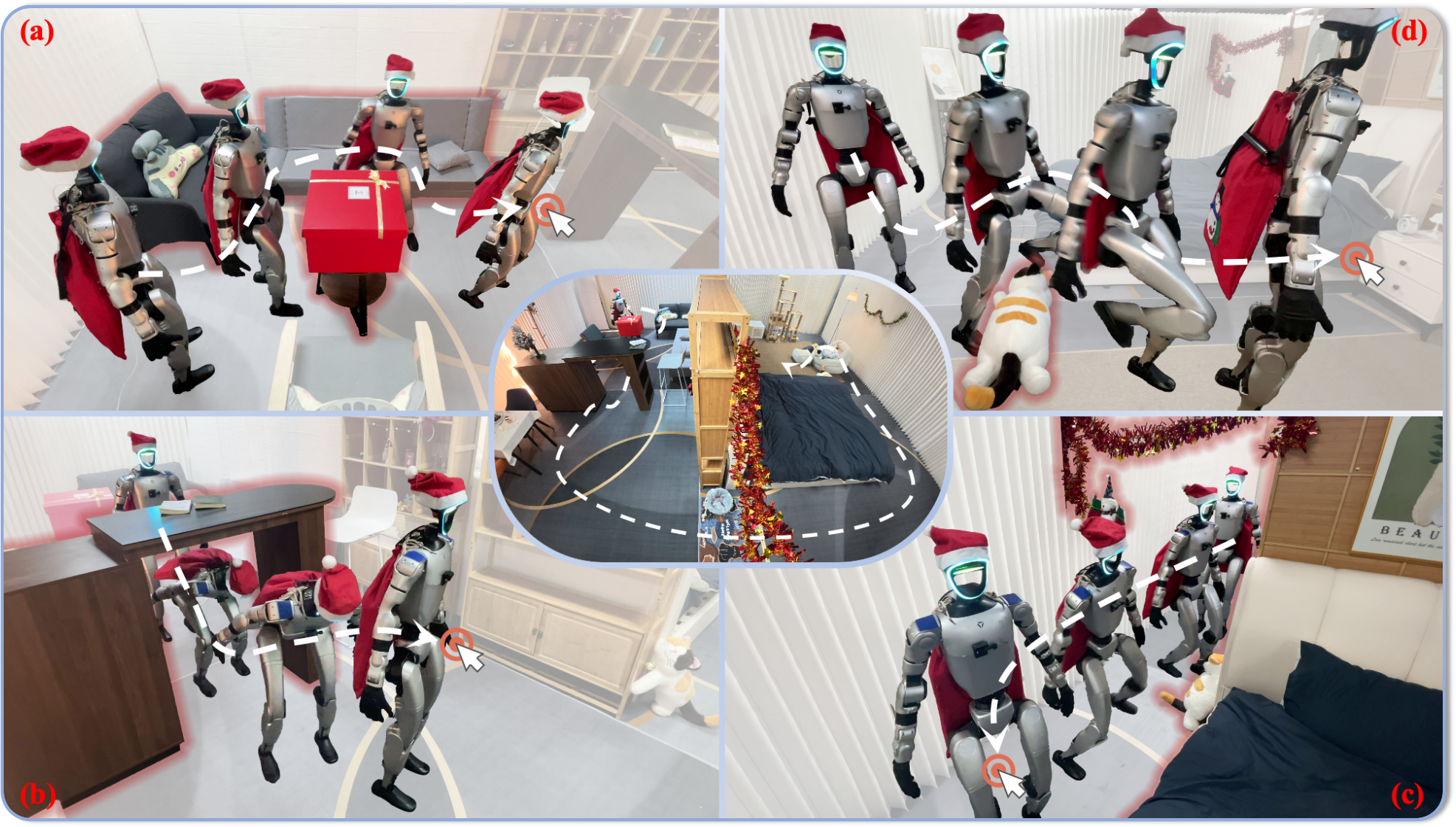}
  \captionof{figure}{Using a single generalist policy, our humanoid robot achieves collision-free traversal in cluttered indoor environments, including {\color{red}(a)} detouring through narrow passages, {\color{red}(b)} crouching under low-hanging obstacles, {\color{red}(c)} squeezing through tight indoor spaces, and {\color{red}(d)} hurdling over objects scattered on the floor.
  }
  \label{fig:teaser}
}
\makeatother

\title{\LARGE \bf
Collision-Free Humanoid Traversal in Cluttered Indoor Scenes
}

\author{Han Xue$^{1,3*}$, Sikai Liang$^{2,3*}$, Zhikai Zhang$^{1,3*}$, Zicheng Zeng$^{3,5}$, Yun Liu$^{1,3}$, Yunrui Lian$^{1,3}$, Jilong Wang$^{3,6}$, \\
Qingtao Liu$^{3,7}$, Xuesong Shi$^{3}$, and Li Yi$^{\dagger,1,4}$
\thanks{*Equal Contributions, $\dagger$Corresponding Author}
\thanks{$^{1}$Tsinghua University, $^{2}$Tongji University, $^{3}$Galbot, $^{4}$Shanghai Qi Zhi Institute, $^{5}$South China University of Technology, $^{6}$Peking University, $^{7}$Zhejiang University}
}

\maketitle
\vspace{-2.0em}
\begin{abstract}

We study the problem of collision-free humanoid traversal in cluttered indoor scenes, such as hurdling over objects scattered on the floor, crouching under low-hanging obstacles, or squeezing through narrow passages. To achieve this goal, the humanoid needs to map its perception of surrounding obstacles with diverse spatial layouts and geometries to the corresponding traversal skills. 
However, the lack of an effective representation that captures humanoid–obstacle relationships during collision avoidance makes directly learning such mappings difficult. We therefore propose Humanoid Potential Field (HumanoidPF), which encodes these relationships as collision-free motion directions, significantly facilitating RL-based traversal skill learning. We also find that HumanoidPF exhibits a surprisingly negligible sim-to-real gap as a perceptual representation.
To further enable generalizable traversal skills through diverse and challenging cluttered indoor scenes, we further propose a hybrid scene generation method, incorporating crops of realistic 3D indoor scenes and procedurally synthesized obstacles. We successfully transfer our policy to the real world and develop a teleoperation system where users could command the humanoid to traverse in cluttered indoor scenes with just a single click. Extensive experiments are conducted in both simulation and the real world to validate the effectiveness of our method. Demos and code can be found in our website: \href{https://axian12138.github.io/CAT/}{https://axian12138.github.io/CAT/}.

\end{abstract}   
\section{Introduction}
\label{sec:intro}

Consider a domestic humanoid robot needs to frequently traverse between the bedroom, living room, and kitchen to perform household chores. A key challenge for the robot is to avoid collisions with the surrounding environment during movement, preventing potential damage to the robot itself or the environment. In cluttered indoor scenes, the humanoid may need to hurdle over objects scattered on the floor, crouch under low-hanging obstacles, or squeeze through narrow passages. This requires the robot to perceive the environment and map obstacles with diverse spatial layouts and geometries to the corresponding traversal skills.

While legged locomotion in complex environments has seen remarkable advances for quadrupeds~\cite{grandia2023perceptive, lee2020learning, gagne2025acrobotics, zhang2024learning, liao2023walking, lee2024learning, yang2023neural, rudin2022advanced, wang2025omni, hoeller2024anymal, miki2024learning, zhuang2023robot, rudin2025parkour, chen2025learning, he2024agile, cheng2024extreme, miki2022learning} and humanoids~\cite{li2023autonomous, he2025attention, sun2025dpl, duan2024learning, zhuang2024humanoid, long2025learning, ren2025vb, huang2025traversing, wang2025beamdojo, sun2025learning, allshire2025visual, ben2025gallant, zhang2025track}, existing works are often limited in their ability to handle traversal in \textit{cluttered indoor scenes} (full-spatial obstacle layouts and intricate, realistic geometries), as shown in Table~\ref{table:comparison}. 
These limitations collectively point to the lack of an effective representation for humanoid–obstacle relationships during collision avoidance: 
(i) existing works~\cite{cheng2024extreme, miki2022learning, hoeller2024anymal, he2024agile, zhuang2024humanoid, ben2025gallant, wang2025omni, miki2024learning, zhuang2023robot, rudin2025parkour, chen2025learning} typically obtain penalty signals only when collisions occur, yielding sparse and delayed supervision. This forces reinforcement learning (RL) to depend on inefficient trial-and-error exploration, thus calling for a representation that can provide anticipatory and dense guidance;
(ii) conventional representations expose the policy to raw, high-dimensional environmental measurement independently of humanoid–obstacle spatial relationships, forcing the policy to infer traversal decisions through implicit kinematic reasoning.

To bridge these gaps, we introduce Humanoid Potential Field (\textbf{HumanoidPF}), an informative representation that encodes humanoid–obstacle relationships for collision avoidance. Inspired by classical Artificial Potential Fields (APF)~\cite{khatib1986real}, HumanoidPF models how the humanoid is influenced by and should react to its surrounding environment as a continuous and differentiable gradient field, inducing ``virtual forces" that point toward collision-free motion directions.

We seamlessly integrate HumanoidPF into traversal skill learning in two complementary ways.
First, HumanoidPF serves as the policy observation by being queried at multiple key body parts, providing directional cues that indicate how each part should move to avoid obstacles and advance toward the goal. This allows the policy to reason directly over traversal decisions, instead of inferring collision-avoidance behavior from raw, high-dimensional visual inputs.
Second, HumanoidPF streamlines collision-aware reward design. The field induces a distribution over preferred motion directions, and the policy is encouraged to align its motion with this distribution. This provides anticipatory and sufficient supervision for RL models, while exhibiting strong cross-scene generalization without manual reward tuning.
Moreover, we observe that HumanoidPF yields a surprisingly negligible sim-to-real gap as a perceptual representation. Its continuous field formulation naturally functions as a low-pass perceptual filter, smoothing out isolated perception artifacts and promoting robust sim-to-real transfer.

To learn traversal skills with HumanoidPF across diverse and challenging obstacle configurations, we propose a hybrid scene generation strategy that systematically expands the space of training scenarios. By augmenting crops of realistic 3D indoor datasets with procedurally synthesized highly constrained obstacles, we expose the robot to a curriculum of challenging clutter configurations that are rarely present in existing datasets, enabling it to acquire rich collision-avoidance experience and substantially improving robustness in near-collision and emergency scenarios.

We further instantiate our approach into a practical teleoperation system, termed Click-and-Traverse (CAT), where the user can simply click a goal to command the humanoid to safely traverse cluttered indoor environments. Extensive experiments in both simulation and realistic real-world indoor scenes validate the practical applicability of HumanoidPF and its strong generalization across diverse environments.

Our contributions are fourfold:
\begin{itemize}
    \item To the best of our knowledge, we are the first to systematically study collision-free humanoid traversal in \textit{cluttered indoor scenes}, advancing toward real-world domestic humanoid robot application.
    \item We propose \textbf{HumanoidPF}, an informative representation that explicitly encodes humanoid–obstacle relationships for collision avoidance, thus significantly facilitating RL-based traversal skill learning.
    \item We propose a hybrid scene generation strategy that exposes the policy to realistic, diverse and challenging cluttered scenarios, significantly improving robustness and generalization in complex indoor environments.
    \item We successfully transfer our policy to the real world as a convenient and useful teleoperation system to command the humanoid to traverse in cluttered indoor scenes.
\end{itemize}

\begin{table}[]
\resizebox{\columnwidth}{!}{
\begin{tabular}{ccc}
\toprule
Method & Spatial layouts & Intricate geometries \\ 
\midrule
PIM~\cite{long2025learning} &  $S = \{g\}$ & \ding{55}\\
HumanoidParkour~\cite{zhuang2024humanoid}  & $S = \{g\}$ & \ding{55} \\
BeamDojo~\cite{wang2025beamdojo} & $S = \{g\}$ & \ding{55} \\
Vb-com~\cite{ren2025vb}& $S \subset \{g,l\},\ |S|=1$ & \ding{55} \\
Gallant~\cite{ben2025gallant}  & $S \subset \{g,l,o\},\ |S|=1$ &  \ding{55}  \\
\midrule
\textbf{Ours}  & $S = \{g,l,o\}$ & \ding{51} \\ 
\bottomrule
\end{tabular}}

\caption{\textbf{Overall comparison with existing works.} $S$: Spatial layouts. $g,l,o$: Ground, lateral and overhead obstacles.}
\label{table:comparison}
\vspace{-1.0em}
\end{table}
\section{Related Works}
\label{sec:relatedwork}

\begin{figure*}
    \centering
    \includegraphics[width=\linewidth]{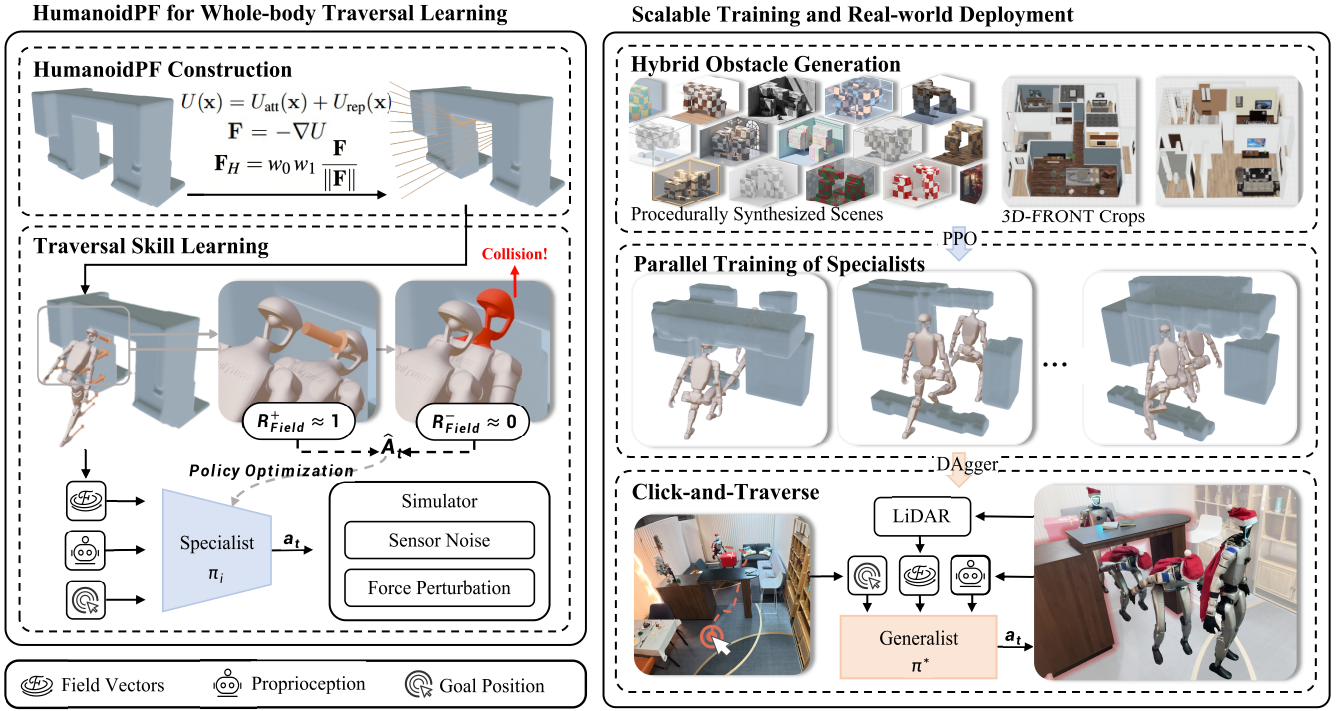}
    \caption{\textbf{Overall pipeline.} 
    We learn a visuomotor policy that maps diverse obstacle geometries and spatial layouts to corresponding whole-body traversal skills.  
    \textbf{Left: HumanoidPF for whole-body traversal learning.}
    \textit{(Top)} Construction of HumanoidPF, a reformulation of APF tailored for humanoid whole-body traversal; 
    \textit{(Bottom)} its use as informative perceptual representation and collision-avoidance rewards.
    \textbf{Right: Scalable training and deployment pipeline.}
    \textit{(Top)} Hybrid scene generation for constructing diverse and challenging training environments; 
    \textit{(Middle)} parallel training of multiple specialist policies followed by distillation into a single generalist policy;
    \textit{(Bottom)} sim-to-real deployment via ``Click-and-Traverse", an intuitive loco-navigation teleoperation in cluttered indoor scenes.
    Sections~\ref{sec:apf_humanoid}, ~\ref{sec:generalize} and ~\ref{sec:click_and_traverse} provide detailed descriptions of the HumanoidPF for traversal learning, the scalable training, and deployment pipeline, respectively.}
    \label{fig:pipeline}
\vspace{-1.8em}
\end{figure*}

\subsection{Legged locomotion in complex environments} 
Legged robots are expected to perform stable locomotion in complex environments, including challenging terrains and obstacles. Quadruped robots have demonstrated robust parkour capabilities on highly challenging terrains~\cite{grandia2023perceptive, lee2020learning, zhang2024learning, cheng2024extreme, lee2024learning, miki2022learning, rudin2022advanced} and confined or cluttered spaces~\cite{wang2025omni, hoeller2024anymal, miki2024learning, zhuang2023robot, rudin2025parkour, chen2025learning}. 
Humanoids have also demonstrated the ability to navigate in height-constrained environments ~\cite{li2023autonomous} and advanced locomotion skills against risky terrains or obstacles, such as stepping stairs, balance beams, and stepping stones ~\cite{zhuang2024humanoid, he2025attention, long2025learning, sun2025dpl, ren2025vb, huang2025traversing, wang2025beamdojo, sun2025learning, ben2025gallant, zhang2025track}.

However, existing works on humanoids often limited to obstacles with partial spatial layouts (e.g. terrains~\cite{zhuang2024humanoid, he2025attention, long2025learning, sun2025dpl, huang2025traversing, wang2025beamdojo, sun2025learning, ren2025vb, zhang2025track}, or over-hanging obstacles~\cite{li2023autonomous}) and simple geometries (e.g. rectangular blocks~\cite{zhuang2024humanoid, li2023autonomous, he2025attention, huang2025traversing, wang2025beamdojo, sun2025learning, sun2025dpl}, or regular polyhedra~\cite{ben2025gallant, ren2025vb}). 
Notably, while Gallant~\cite{ben2025gallant} addresses ground, lateral, and overhead obstacle layouts in isolation, it does not consider scenarios where these constraints coexist.
In contrast, our method constructs HumanoidPF to operate in \textit{cluttered indoor scenes} where full-spatial constraints are jointly present with highly intricate geometries. The comparison of existing works and our work is shown in Table~\ref{table:comparison}.

\subsection{Artificial potential field for obstacle avoidance}
Originally introduced in the late 1980s, the Artificial Potential Field (APF)~\cite{khatib1986real} method generates a virtual force field to guide the motion of manipulators or mobile robots for obstacle avoidance. Inspired by physical analogies, the goal position is modeled as an attractive pole, while obstacles act as repulsive surfaces. 
Traditionally, APF has been widely used in 2D path planning for mobile robots ~\cite{wu2023robot, hwang1992potential, ge2002dynamic} and robotic manipulators ~\cite{gong2025geopf, ginesi2021dynamic, chen2023research}.

However, only a few studies have combined model-based quadruped control with APF in limited forms by abstracting the center of mass~\cite{igarashi2006free, igarashi2004path} or foot joint~\cite{zhuo2021variable, voeurn2021motion} as a single rigid body, which is insufficient to handle the complex planning and control challenges of humanoid learning. In contrast, we propose \textbf{HumanoidPF}, a principled reformulation of APF specifically tailored for informative perception and reward streamlining of humanoid skill learning.

\section{Method}
\label{sec:method}

We study the problem of collision-free humanoid traversal in cluttered indoor scenes. Given a target position $\mathbf{g} \in \mathbb{R}^3$, and a set of indoor obstacles $\mathcal{O} = \{ O_i \}_{i=1}^{N}$, the humanoid needs to move to $\mathbf{g}$ without any collision with $\mathcal{O}$. To solve this problem, the humanoid needs to map its perception of surrounding obstacles to the corresponding traversal skills. Our method can be split into two parts. We first introduce how our \textbf{HumanoidPF} encodes humanoid–obstacle relationships to facilitate humanoid traversal learning in Section~\ref{sec:apf_humanoid}. We further introduce how to generalize our policy to diverse and challenging indoor scenes with our proposed hybrid scene generation method in Section~\ref{sec:generalize}. 
For real-world deployment, we further instantiate our approach as a teleoperated loco-navigation system, which is presented in Section~\ref{sec:click_and_traverse}. The overall pipeline is shown in Figure~\ref{fig:pipeline}. 

\subsection{HumanoidPF for whole-body traversal learning}
\label{sec:apf_humanoid}
We substantially extend classical APF to support learning-based whole-body humanoid traversal. In APF, the target location $\mathbf{g}$ is modeled as an attractive pole and obstacles $\mathcal{O}$ as repulsive surfaces, forming a gradient field that indicates collision-free motion toward the goal. However, prior works directly apply APF in single-rigid-body model-based control, which is inadequate for the high-dimensional and tightly coupled planning and control demands of humanoid skill learning. We therefore propose HumanoidPF, a principled reformulation of APF tailored for humanoid, which encodes humanoid–obstacle relationships for informative perception and reward streamlining.

\subsubsection{HumanoidPF construction}
\label{sec:humanoidpf_construction}
We begin by constructing the attractive field $U_{\text{att}}$:
\begin{equation}
U_{\text{att}}(\mathbf{x}) = \eta \, \| \mathbf{x} - \mathbf{g} \|_{\text{geo}},
\end{equation}
where the geodesic distance $\| \mathbf{x} - \mathbf{g} \|_{\text{geo}}$ represents the shortest 3D path from position $\mathbf{x}$ to the goal $\mathbf{g}$ without intersecting obstacles, and $\eta$ is a scaling factor. The geodesic distance inherently accounts for obstacle geometry, providing safer guidance than a simple Euclidean distance.

Next, the repulsive field $U_{\text{rep}}$ prevents collisions and is defined as:
\begin{equation}
U_{\text{rep}}(\mathbf{x}) =
\begin{cases}
\frac{1}{2}\,\xi\,\left(\frac{1}{d(\mathbf{x})} - \frac{1}{d_0}\right)^2, & d(\mathbf{x}) \le d_0, \\[6pt]
0, & d(\mathbf{x}) > d_0,
\end{cases}
\end{equation}
where $d(\mathbf{x})$ is the signed distance, $\xi$ is a scaling factor, and $d_0$ defines the influence range of obstacles. 

\begin{figure}[t]
  \centering
   \includegraphics[width=0.95\linewidth]{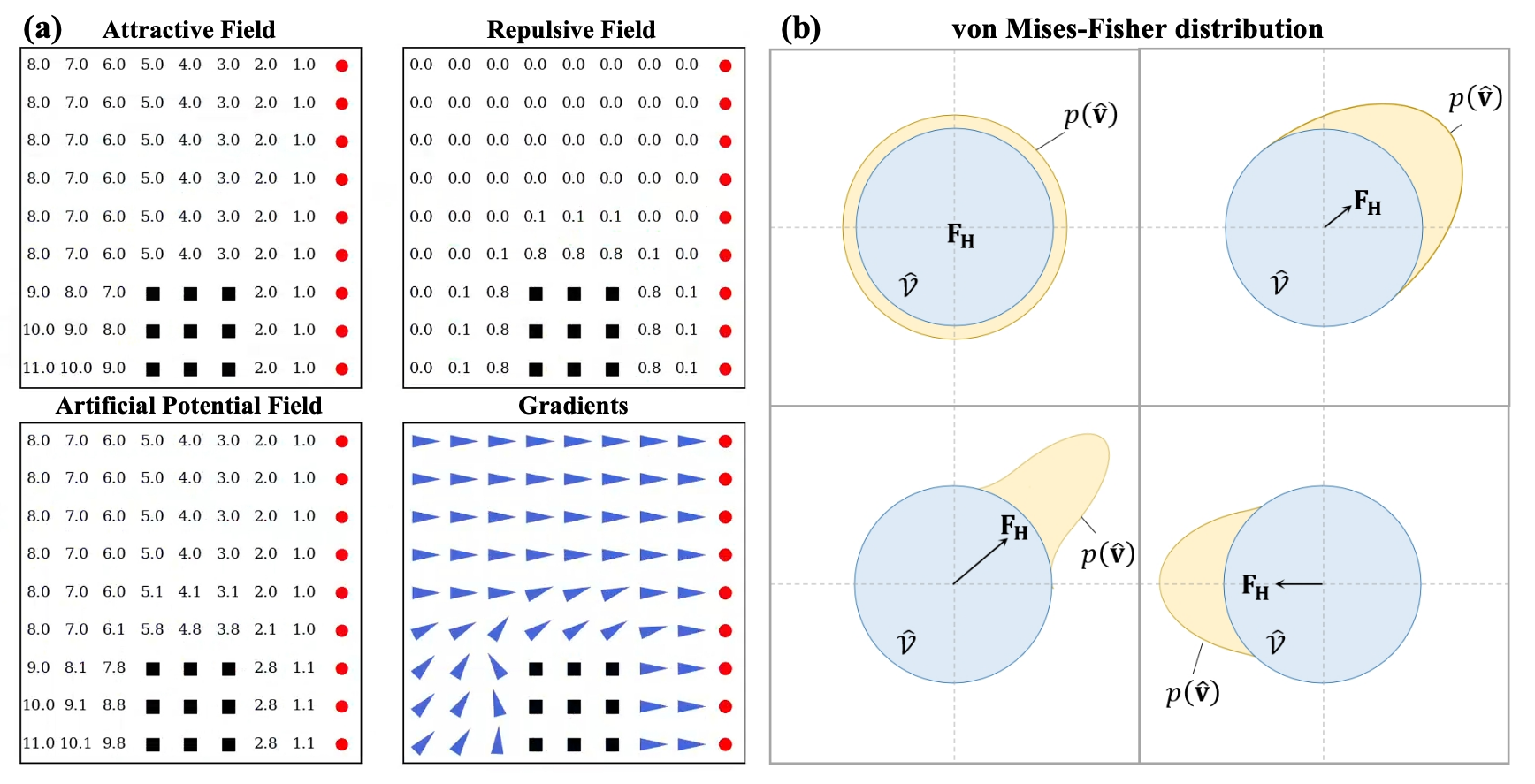}
   \caption{(a) Construction of the APF and (b) motion prior distribution induced by the HumanoidPF. }
   \label{fig:field-vMF}
\vspace{-1.8em}
\end{figure}

The final guidance field is the negative gradient of a combined potential,
\begin{equation}
\mathbf{F} = -\nabla U,\quad
U(\mathbf{x}) = U_{\text{att}}(\mathbf{x}) + U_{\text{rep}}(\mathbf{x}),
\label{eq:field_direction}
\end{equation}
which is then queried at the locations of different body parts, yielding field vectors $\mathbf{F}(\mathbf{x}_k)$ for each body part $\mathbf{p}_k$. A 2D visualization of our APF is illustrated in Figure.~\ref{fig:field-vMF} (a).

While APF method typically models robots as a single rigid body, its direct application to multi-jointed humanoid robots could lead to conflicts between body parts. For instance, when the robot faces an obstacle directly ahead, it must decide whether to move left or right. The potential fields on the left and right sides of the body direct it toward opposite paths. In a symmetrical configuration, these vectors cancel each other out, leading to a multi-modal dilemma where the robot becomes trapped in a local minimum or exhibits oscillatory behavior. To address this challenge, we propose a priority-weighting scheme that prioritizes the influence of certain body parts over others according to their contribution to the task.

\textbf{Priority-weighting.} Instead of treating all body parts equally, our priority-weighting scheme adjusts the influence of each body part based on its role in the overall motion.

To establish coherent global guidance, we assign a higher priority to the root body part (e.g., pelvis) since it plays a central role in maintaining stability and direction:
\begin{equation}
w_0(\mathbf{p}_{\text{root}}) = 1,\quad 
w_0(\mathbf{p}_{\text{others}}) = 0.5.
\end{equation}

Furthermore, some body parts are more critical in avoiding obstacles, particularly those closer to potential collisions. To account for this, we define a dynamic collision-urgency weight based on the signed distance $d(\mathbf{x}_k)$ and the Cartesian velocity $\mathbf{v}_k$ of body part $\mathbf{p}_k$, with a scaling factor $\lambda$:
\begin{equation}
w_1(\mathbf{p}_k) = \lambda \,
\mathrm{max}\!\left( - \nabla d(\mathbf{x}_k) \cdot \mathbf{v}_k,\; 0.5 \right)\exp\!\big(-d(\mathbf{x}_k)\big).
\label{eq:field_magnitude}
\end{equation}

The resulting HumanoidPF is defined as $
\mathbf{F}_H = w_0\, w_1 \,\frac{\mathbf{F}}{\lVert \mathbf{F} \rVert}$. 
This scheme attenuates conflicting influences and promotes coordinated whole-body control. In particular, minute asymmetries in the spatial configuration are selectively amplified, thus seamlessly resolving the multi-modal dilemma.

\subsubsection{Traversal skill learning with HumanoidPF}

\textbf{HumanoidPF for policy observation.}
To better inform RL policies about humanoid-obstacle relationships, we leverage HumanoidPF to construct a compact, task-relevant visual observation. It is sampled at $K=13$ body parts,
\begin{equation}
OBS_{Field} = \{ \mathbf{F}_H(\mathbf{x}_k) \mid \mathbf{x}_k\}^K_{k = 1},
\end{equation}
where each $\mathbf{F}_H(\mathbf{x}_k)$ encodes the local directional guidance induced by obstacles and the target at body part $k$, indicating collision-free motion. Sampling these fields at key body parts specifies how the humanoid should steer its body through the environment, allowing the policy to reason about traversal decisions rather than implicit inference from raw visual data. We empirically validate this in Section~\ref{sec:apf_for_humanoid}.

HumanoidPF for observation further mitigates the perceptual sim-to-real gap by representing the environment as a continuous, spatially aggregated field, which functions like a low-pass perceptual filter. Unlike raw sensor representations that retain fine-grained geometric details and are sensitive to small local perturbations, the field formulation suppresses isolated noise while preserving the dominant spatial gradients relevant to the traversal task. This ensures that minor geometric variations do not significantly affect control during real-world deployment, as is empirically validated in Section~\ref{sec:sim2real_transfer}

\begin{figure*}
    \centering
    \includegraphics[width=\linewidth]{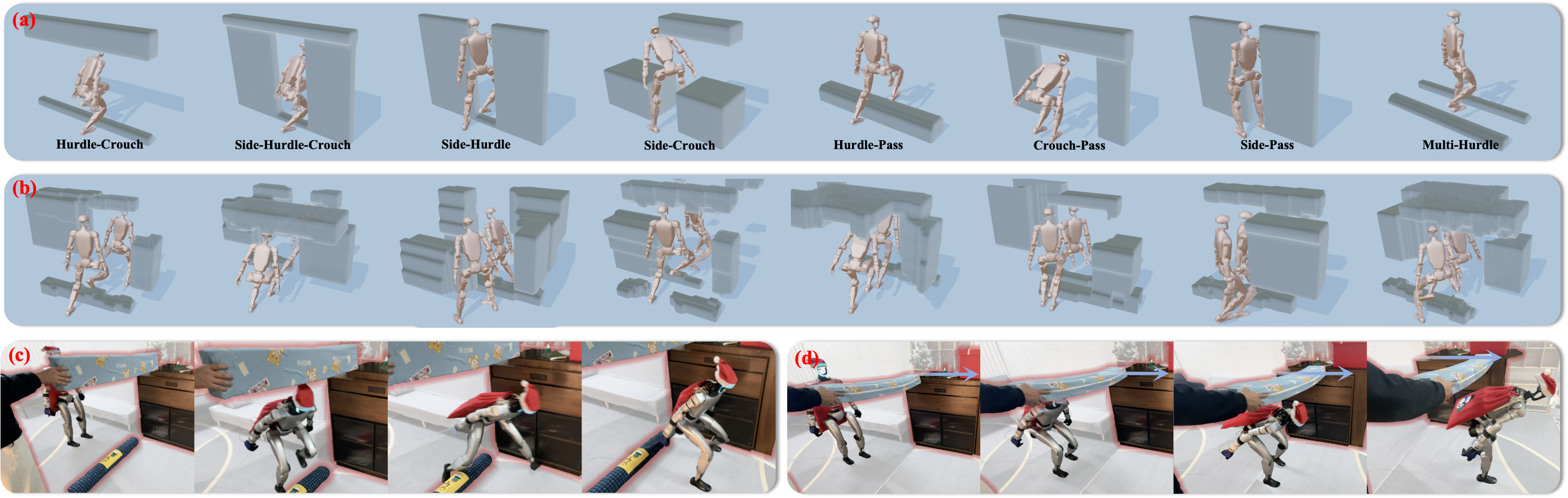}
    \caption{\textbf{Collision-free humanoid traversal in both simulation and the real world.} {\color{red}(a)} Humanoid traversal behaviors on eight representative test scene types; {\color{red}(b)} traversal behaviors in procedurally generated cluttered environments; {\color{red}(c)} real-world ``hurdle-crouch" scenario, validating sim-to-real transfer in a cluttered indoor setting; {\color{red}(d)} robustness under dynamic disturbances, where simple object movements ({\color{lightblue}blue arrows}) are introduced during traversal.
    }
\label{fig:more_visualization}
\vspace{-1.5em}
\end{figure*}

\textbf{HumanoidPF for policy reward.}
To streamline reward engineering, we employ HumanoidPF to induce anticipatory and dense guidance that generalizes across diverse environments. At each time step, HumanoidPF encodes a distribution of preferred motion directions, and the policy is optimized to produce actions that align with this distribution, thus fostering safe and dexterous collision-avoidance behaviors.

The von Mises–Fisher (vMF) distribution is used to model directional preferences $\boldsymbol{\mu}(\mathbf{x}) \in \mathbb{R}^3$ on the unit sphere and allows the strength of this preference to be controlled by a single concentration parameter $\kappa(\mathbf{x}) \in \mathbb{R}$:
\begin{equation}
p(\hat{\mathbf{v}} \mid \boldsymbol{\mu}(\mathbf{x}), \kappa(\mathbf{x}))
= C_d(\kappa)\exp\!\big(\kappa(\mathbf{x})\, \boldsymbol{\mu}(\mathbf{x})^\top \cdot \hat{\mathbf{v}}\big),
\end{equation}
where $\hat{\mathbf{v}}$ is the motion direction of a humanoid body part, $C_d(\kappa)$ is the normalization function.  

$\boldsymbol{\mu}(\mathbf{x})$ and $\kappa(\mathbf{x})$ is directly derived from HumanoidPF:
\begin{equation}
\boldsymbol{\mu}(\mathbf{x}) = \frac{\mathbf{F}_H(\mathbf{x})}{\lVert \mathbf{F}_H(\mathbf{x}) \rVert}, \quad
\kappa(\mathbf{x}) = \kappa_{\max} \lVert \mathbf{F}_H(\mathbf{x}) \rVert,
\end{equation}
where $\kappa_{\max}$ is a scaling factor. The body part with higher priority receives a field vector $\mathbf{F}_H(\mathbf{x})$ with larger magnitude; accordingly, $\kappa(\mathbf{x})$ increases to enforce stricter alignment with $\boldsymbol{\mu}(\mathbf{x})$ , and vice versa for lower-priority parts. This priority-aware concentration design promotes coordinated whole-body motion while improving collision-avoidance behavior, as illustrated in Figure~\ref{fig:field-vMF} (b).

During policy training, let the motion direction of the $k$-th body part be $\hat{\mathbf{v}}_k = \mathbf{v}_k / \|\mathbf{v}_k\|$, the associated prior direction be $\boldsymbol{\mu}_k = \boldsymbol{\mu}(\mathbf{x}_k)$ and concentration parameter be $\kappa_k = \kappa(\mathbf{x}_k)$. 
Assuming independence among joints, the whole-body motion prior and log-likelihood reward are expressed as:
\begin{equation}
p(\hat{\mathbf{v}}_{1:K} \mid \mathbf{x}_{1:K}, \mathbf{F}_H)
= \prod_{k=1}^{K}
C_d(\kappa_k)\,
\exp\!\big(\kappa_k\,\boldsymbol{\mu}_k^\top \cdot \hat{\mathbf{v}}_k\big),
\end{equation}

\begin{equation}
R_{Field}
= \sum_{k=1}^{K}
\Big[
\log C_d(\kappa_k) + \kappa_k\,\boldsymbol{\mu}_k^\top \cdot \hat{\mathbf{v}}_k
\Big].
\end{equation}

This reward formulation exhibits strong cross-scene generalization without requiring manual tuning, thereby enabling an automated training pipeline that scales effectively across diverse environments.

\subsection{Scalable training in diverse and challenging scenes}
\label{sec:generalize}
For general practical use, the humanoid needs to handle diverse scenes within a single unified policy. It needs the policy trained with a sufficiently large and challenging indoor scene dataset to enable generalization in real-world cluttered scenes. Therefore, we propose a  hybrid scene generation method in Section~\ref{sec:hybrid}. It incorporates crops of realistic 3D indoor scenes for structural realism, and procedurally synthesized obstacles to enrich highly challenging clutter configurations. 
In addition, even with HumanoidPF, directly learning a single policy across all scenes remains challenging due to the low sample efficiency of RL. We therefore adopt a specialist-to-generalist training strategy inspired by~\cite{ross2011reduction, zhang2025unleashing}, described in Section~\ref{sec:spe_to_gen}.

\subsubsection{Hybrid scene generation}
\label{sec:hybrid}
We observe that highly challenging obstacle layouts constitute merely a long-tail subset in most existing datasets~\cite{fu20213d, deitke2022️, ramakrishnan2021habitat}, since typical indoor scenes feature orderly object arrangements and clearly delineated walkable regions. 
Simply scaling up the dataset does not alleviate this problem. Therefore, we propose a novel hybrid scene generation scheme that enriches realistic 3D indoor datasets with procedurally synthesized "extreme" obstacles, where full-spatial constraints are jointly present.

\textbf{Crops of realistic 3D indoor scenes.} For generalization to intricate and realistic indoor environments, we adopt the 3D-FRONT~\cite{fu20213d} dataset, containing structurally realistic scenes with large-scale high-quality furniture objects. We selectively crop and filter scene blocks for policy training.

Specifically, we first project all furniture onto the ground and erode the resulting planar walkable regions with a radius of 0.1 m to account for clearance. Within the remaining walkable regions, we randomly sample a start location and crop a 5 m × 5 m block centered at this position. During training, the goal location will be randomly sampled on a circle with a radius of 2 m around the start.

We initially train specialist policies on all such cropped scenes and subsequently identify scenes with low traversal success rates. Scenes that are empirically found to be non-traversable are manually filtered out.

\begin{table*}[!t]
    \centering
    \renewcommand{\arraystretch}{1.03} 
    \setlength{\tabcolsep}{4pt} 
    \begin{tabular*}{0.95\textwidth}{@{\extracolsep{\fill}} l  llll llll }
        \toprule
        \multicolumn{1}{c}{\textbf{}}  
        & \multicolumn{2}{c}{\textbf{Hurdle-Crouch}} 
        & \multicolumn{2}{c}{\textbf{Side-Hurdle-Crouch}}
        & \multicolumn{2}{c}{\textbf{Side-Hurdle}} 
        & \multicolumn{2}{c}{\textbf{Side-Crouch}} 
        \\
         \cmidrule(lr){2-3} \cmidrule(lr){4-5} 
         \cmidrule(lr){6-7} \cmidrule(lr){8-9}
         \textbf{} 
         & {\textit{SR(\%)} $\uparrow$} & {\textit{DE(m)} $\downarrow$}
         & {\textit{SR(\%)} $\uparrow$} & {\textit{DE(m)} $\downarrow$} 
         & {\textit{SR(\%)} $\uparrow$} & {\textit{DE(m)} $\downarrow$} 
         & {\textit{SR(\%)} $\uparrow$} & {\textit{DE(m)} $\downarrow$}  
         \\
        \midrule
        ASTraversal
        & 28.1{\textcolor{gray}{\scriptsize{} $\pm$10.4}} & 1.11{\textcolor{gray}{\scriptsize{} $\pm$0.78}}
        & 0.5{\textcolor{gray}{\scriptsize{} $\pm$0.5}} & 1.06{\textcolor{gray}{\scriptsize{} $\pm$0.39}} 
        & 37.1{\textcolor{gray}{\scriptsize{} $\pm$3.1}} & 0.54{\textcolor{gray}{\scriptsize{} $\pm$0.32}}
        & 56.0{\textcolor{gray}{\scriptsize{} $\pm$9.9}} & 0.48{\textcolor{gray}{\scriptsize{} $\pm$0.05}}
        \\
        Humanoid Parkour 
        & 33.3{\textcolor{gray}{\scriptsize{} $\pm$6.1}} & 1.16{\textcolor{gray}{\scriptsize{} $\pm$0.63}}
        & 0.4{\textcolor{gray}{\scriptsize{} $\pm$0.3}} & 1.49{\textcolor{gray}{\scriptsize{} $\pm$0.04}} 
        & 45.1{\textcolor{gray}{\scriptsize{} $\pm$4.5}} & 0.62{\textcolor{gray}{\scriptsize{} $\pm$0.39}}
        & 64.4{\textcolor{gray}{\scriptsize{} $\pm$19.3}} & 0.56{\textcolor{gray}{\scriptsize{} $\pm$0.16}}
        \\
        \arrayrulecolor{black!50}
        \midrule
        \arrayrulecolor{black}
        Ours w/o $OBS_{Field}$ 
        & \second{77.8{\textcolor{gray}{\scriptsize{} $\pm$5.4}}} & \second{0.33{\textcolor{gray}{\scriptsize{} $\pm$0.23}}} 
        & \second{53.7{\textcolor{gray}{\scriptsize{} $\pm$9.9}}} & \second{0.59{\textcolor{gray}{\scriptsize{} $\pm$0.08}}} 
        & 60.4{\textcolor{gray}{\scriptsize{} $\pm$9.6}} & 0.53{\textcolor{gray}{\scriptsize{} $\pm$0.68}} 
        & \second{90.1{\textcolor{gray}{\scriptsize{} $\pm$5.3}}} & \second{0.19{\textcolor{gray}{\scriptsize{} $\pm$0.35}}} 
        \\
        Ours w/o $R_{Field}$ 
        & 21.9{\textcolor{gray}{\scriptsize{} $\pm$15.8}} & 1.27{\textcolor{gray}{\scriptsize{} $\pm$0.71}} 
        & 0.0{\textcolor{gray}{\scriptsize{} $\pm$0.0}} & 1.57{\textcolor{gray}{\scriptsize{} $\pm$0.003}} 
        & \second{71.4{\textcolor{gray}{\scriptsize{} $\pm$9.9}}} & \second{0.5{\textcolor{gray}{\scriptsize{} $\pm$0.34}}} 
        & 80.3{\textcolor{gray}{\scriptsize{} $\pm$15.4}} & 0.23{\textcolor{gray}{\scriptsize{} $\pm$0.06}} 
        \\
        Ours 
        & \best{93.9{\textcolor{gray}{\scriptsize{} $\pm$2.7}}} & \best{0.08{\textcolor{gray}{\scriptsize{} $\pm$0.16}}} 
        & \best{86.6{\textcolor{gray}{\scriptsize{} $\pm$5.2}}} & \best{0.2{\textcolor{gray}{\scriptsize{} $\pm$0.32}}} 
        & \best{95.4{\textcolor{gray}{\scriptsize{} $\pm$3.9}}} & \best{0.06{\textcolor{gray}{\scriptsize{} $\pm$0.34}}} 
        & \best{96.9{\textcolor{gray}{\scriptsize{} $\pm$2.1}}} & \best{0.05{\textcolor{gray}{\scriptsize{} $\pm$0.09}}} 
        \\
        \midrule
        \multicolumn{1}{c}{\textbf{}}  
        & \multicolumn{2}{c}{\textbf{Hurdle-Pass}} 
        & \multicolumn{2}{c}{\textbf{Crouch-Pass}} 
        & \multicolumn{2}{c}{\textbf{Side-Pass}} 
        & \multicolumn{2}{c}{\textbf{Multi-Hurdle}}
        \\
         \cmidrule(lr){2-3} \cmidrule(lr){4-5} 
         \cmidrule(lr){6-7} \cmidrule(lr){8-9}
         \textbf{} 
         & {\textit{SR(\%)} $\uparrow$} & {\textit{DE(m)} $\downarrow$}
         & {\textit{SR(\%)} $\uparrow$} & {\textit{DE(m)} $\downarrow$} 
         & {\textit{SR(\%)} $\uparrow$} & {\textit{DE(m)} $\downarrow$} 
         & {\textit{SR(\%)} $\uparrow$} & {\textit{DE(m)} $\downarrow$}  
         \\
        \midrule
        ASTraversal
        & 75.9{\textcolor{gray}{\scriptsize{} $\pm$6.8}} & 0.66{\textcolor{gray}{\scriptsize{} $\pm$0.3}}
        & 41.3{\textcolor{gray}{\scriptsize{} $\pm$5.3}} & 0.9{\textcolor{gray}{\scriptsize{} $\pm$1.04}}
        & 55.2{\textcolor{gray}{\scriptsize{} $\pm$8.5}} & 0.78{\textcolor{gray}{\scriptsize{} $\pm$0.87}}
        & 82.1{\textcolor{gray}{\scriptsize{} $\pm$8.7}} & 0.26{\textcolor{gray}{\scriptsize{} $\pm$0.43}} 
        \\
        Humanoid Parkour 
        & 84.3{\textcolor{gray}{\scriptsize{} $\pm$8.0}} & 0.32{\textcolor{gray}{\scriptsize{} $\pm$0.18}} 
        & 48.1{\textcolor{gray}{\scriptsize{} $\pm$3.3}} & 1.34{\textcolor{gray}{\scriptsize{} $\pm$0.18}} 
        & 41.3{\textcolor{gray}{\scriptsize{} $\pm$2.5}} & 0.91{\textcolor{gray}{\scriptsize{} $\pm$0.34}} 
        & 88.7{\textcolor{gray}{\scriptsize{} $\pm$2.6}} & 0.23{\textcolor{gray}{\scriptsize{} $\pm$0.35}} 
        \\
        \arrayrulecolor{black!50}
        \midrule
        \arrayrulecolor{black}
        Ours w/o $OBS_{Field}$ 
        & \second{92.3{\textcolor{gray}{\scriptsize{} $\pm$4.9}}} & \second{0.1{\textcolor{gray}{\scriptsize{} $\pm$0.15}}} 
        & \second{96.8{\textcolor{gray}{\scriptsize{} $\pm$5.0}}} & \second{0.07{\textcolor{gray}{\scriptsize{} $\pm$0.3}}} 
        & \second{95.2{\textcolor{gray}{\scriptsize{} $\pm$4.4}}} & \second{0.07{\textcolor{gray}{\scriptsize{} $\pm$0.22}}} 
        & \second{90.5{\textcolor{gray}{\scriptsize{} $\pm$3.5}}} & \second{0.09{\textcolor{gray}{\scriptsize{} $\pm$0.11}}} 
        \\
        Ours w/o $R_{Field}$ 
        & 90.7{\textcolor{gray}{\scriptsize{} $\pm$7.5}} & 0.12{\textcolor{gray}{\scriptsize{} $\pm$0.37}} 
        & 95.9{\textcolor{gray}{\scriptsize{} $\pm$5.0}} & 0.08{\textcolor{gray}{\scriptsize{} $\pm$0.24}} 
        & 28.0{\textcolor{gray}{\scriptsize{} $\pm$18.3}} & 1.07{\textcolor{gray}{\scriptsize{} $\pm$0.56}}
        & 88.3{\textcolor{gray}{\scriptsize{} $\pm$14.1}} & 0.23{\textcolor{gray}{\scriptsize{} $\pm$0.5}} 
        \\
        Ours
        & \best{96.9{\textcolor{gray}{\scriptsize{} $\pm$5.5}}} & \best{0.06{\textcolor{gray}{\scriptsize{} $\pm$0.15}}} 
        & \best{97.5{\textcolor{gray}{\scriptsize{} $\pm$4.3}}} & \best{0.05{\textcolor{gray}{\scriptsize{} $\pm$0.09}}} 
        & \best{97.3{\textcolor{gray}{\scriptsize{} $\pm$3.2}}} & \best{0.04{\textcolor{gray}{\scriptsize{} $\pm$0.15}}} 
        & \best{95.0{\textcolor{gray}{\scriptsize{} $\pm$4.9}}} & \best{0.06{\textcolor{gray}{\scriptsize{} $\pm$0.1}}} 
        \\
        \bottomrule
    \end{tabular*}
    \caption{\textbf{Validation of HumanoidPF for skill learning.} To better characterize the performance of our method under diverse obstacle layouts, we design 8 distinct scene types for evaluation and ablation studies. }
    \label{tab:apf_comparison}
\vspace{-1.5em}
\end{table*}

\textbf{Procedurally generated obstacles.} To supplement crops of 3D-FRONT with more challenging and cluttered environments, we procedurally generate obstacles that impose full-spatial (simultaneous ground, lateral, and overhead) constraints, deliberately targeting highly restrictive scenarios. Specifically, we place boxes with varying positions, dimensions, and orientations that may extend upward from the floor, descend from the ceiling, and be placed in close proximity to form narrow traversal passages.

To break structural regularity and enhance geometric realism, we apply random $\mathrm{SO}(3)$ rotations and 2D Perlin noise to each box. The resulting artifacts, such as spiky surfaces or non-manifold regions, are mitigated via 3D morphological closing and opening at the voxel level before mesh conversion. Visualizations of the robot traversing generated obstacles are shown in Figure~\ref{fig:pipeline} and Figure~\ref{fig:more_visualization} (b).

To support curriculum learning during policy training, we use a layout-agnostic difficulty factor to control obstacle complexity, such as the number and size of boxes. As the difficulty increases, the policy progressively acquires robust traversal skills under increasingly challenging configurations.

\subsubsection{Specialist-to-generalist training}
\label{sec:spe_to_gen}
We construct an automated and parallel specialist-to-generalist training pipeline. We first conduct large-scale training of specialist policies across diverse scenes via PPO~\cite{schulman2017proximal}. The reward derived from HumanoidPF is scene-general, enabling scalable training on 139 cropped 3D-FRONT scenes and 216 procedurally generated scenes. Each specialist is trained with 32,768 parallel environments and 5,000,000 episodes, with start and goal locations randomly sampled for each episode.

Subsequently, a generalist policy is distilled using DAgger~\cite{ross2011reduction} from multiple specialists as teacher policies. Dedicated specialists provide expert actions conditioned on current obstacle, enabling the generalist to acquire strong generalization capability across varying scenarios.

To learn robust and stable traversal skills, both specialist and generalist policies are trained with sensor noise and force perturbations to simulate realistic collision-avoidance conditions. In addition, a curriculum with progressively increasing scene difficulty is employed to gradually enhance traversal performance and explore the limits of the policy’s obstacle-avoidance capability.

\subsection{Real-world deployment: Click-and-Traverse}
\label{sec:click_and_traverse}
For real-world deployment, we instantiate our method as a teleoperated loco-navigation system, termed Click-and-Traverse (CAT). The system integrates a LiDAR–inertial SLAM pipeline based on Fast-LIO2~\cite{xu2021fastlio2fastdirectlidarinertial} and OctoMap~\cite{hornung13auro}. It maintains an up-to-date environment mapping and field construction, both operating at a frequency of 10 Hz. The policy queries the HumanoidPF at different body parts via self-localization and forward-kinematics. 

Users could specify a desired goal by simply clicking on a grid map, after which the humanoid autonomously navigates to the target while avoiding collisions. This interface removes the need for labor-intensive control modalities such as joysticks or motion capture, providing a lightweight and highly automated teleoperation solution.

\section{Experiment}
\label{sec:exp}

In this section, we provide extensive experimental results in both the MuJoCo~\cite{todorov2012mujoco} simulator and the real world on Unitree G1 humanoid robot. The experiments aim to address the following three questions:

\begin{itemize}
    \item \textbf{Q1}: Can \textbf{HumanoidPF} improve the performance of traversal in cluttered indoor scenes compared to existing methods?
    \item \textbf{Q2}: Can our hybrid scene generation method help the policy generalize to unseen and challenging scenes?
    \item \textbf{Q3}: Can the HumanoidPF for observation $OBS_{Field}$ help the sim-to-real transfer?
\end{itemize}

\subsection{Validation of HumanoidPF for skill learning}
\label{sec:apf_for_humanoid}
To address \textbf{Q1} (\textit{Can HumanoidPF improve the performance of traversal in cluttered indoor scenes compared to existing methods?}), we compare the performance of our method against existing ones on traversing cluttered scenes.

\textbf{Experiment Setting.} To systematically analyze performance under different obstacle layouts and geometric configurations, we design eight distinct types of cluttered scenes for evaluation. All scenes used in this experiment are manually generated, each type exhibiting distinct characteristics and collectively covering a broad range of challenging obstacle configurations, with 10 scenes per type. We train and evaluate all methods on these generated scenes to compare their ability to traverse cluttered environments. Representative visualizations of the robot traversing each scene type are shown in Figure~\ref{fig:more_visualization} (a).

\textbf{Experiment Metrics.} We use the following metrics to evaluate the performance on traversing cluttered scenes:
\begin{itemize}
    \item \textbf{Success Rate (SR, \%)} records the percentage of successful traversal trials. A trial is considered successful if the robot reaches within 0.1 meters of the target location in 5 seconds without colliding with any obstacle.
    
    \item \textbf{Distance Error (DE, m)} is the averaged closest horizontal distance between the humanoid root and the target location in a traversal.
    
\end{itemize}

\textbf{Baselines.} We choose the following methods and adapt them to fit our problem setting for a fair comparison:
\begin{itemize}
    \item \textbf{ASTraversal}~\cite{chen2025learning}: ASTraversal was originally proposed for quadrupeds. We re-implement its core multi-layer elevation maps and obstacle-avoidance reward for our humanoid framework.
    
    \item \textbf{Humanoid Parkour}~\cite{zhuang2024humanoid}: The method enhances collision penalties to better guide locomotion, but is limited to terrain modeling below the feet. To adapt it to our full-space traversal setting, we additionally provide overhead obstacle perception for fair comparison.
    
    \item \textbf{Ours w/o $OBS_{Field}$:} We replace the HumanoidPF in observation with multi-layer elevation maps as used in ASTraversal~\cite{chen2025learning}. We adopt this baseline to evaluate the effectiveness of HumanoidPF for observation.

    \item \textbf{Ours w/o $R_{Field}$:} We remove the HumanoidPF-guided reward. Instead, we use a basic collision-penalty reward commonly used in collision-avoidance tasks~\cite{zhuang2024humanoid}. We adopt this baseline to validate HumanoidPF for reward. 
\end{itemize}

\begin{table}[!t]
    \centering
    \renewcommand{\arraystretch}{1.05}
    \setlength{\tabcolsep}{8pt}
    \begin{tabular}{lcc}
        \toprule
        \textbf{Group} & \textbf{Easy (SR\%)} & \textbf{Hard (SR\%)} \\
        \midrule
        Base & 62.0{\textcolor{gray}{\scriptsize{} $\pm$23.4}} &  1.2{\textcolor{gray}{\scriptsize{} $\pm$2.9}} \\
        Base + Syn-Partial-Easy & 78.6{\textcolor{gray}{\scriptsize{} $\pm$12.2}} & 12.3{\textcolor{gray}{\scriptsize{} $\pm$10.3}} \\
        Base + Syn-Full-Easy & 83.1{\textcolor{gray}{\scriptsize{} $\pm$19.0}} & 26.4{\textcolor{gray}{\scriptsize{} $\pm$5.8}} \\
        Base + Syn-Full-Hard & 95.2{\textcolor{gray}{\scriptsize{} $\pm$6.1}} & 66.7{\textcolor{gray}{\scriptsize{} $\pm$17.9}} \\
        \bottomrule
    \end{tabular}
    \caption{\textbf{Validation of hybrid scene generation.} Mean success rate (\%) on easy and hard subsets for the four experiment groups (mean $\pm$ std).}
    \label{tab:succ_easy_hard}
\vspace{-1.8em}
\end{table}

\textbf{Experiment Results.} Results are summarized in Table~\ref{tab:apf_comparison}. Our method consistently achieves the best performance across all types of test cases, demonstrating its strong traversal capability in cluttered scenes. In contrast, both Humanoid Parkour and ASTraversal only achieve competitive performance in relatively simple terrains. Nevertheless, our approach exhibits low variance over multiple trials, indicating high robustness and stability of the learned policy.

\subsection{Validation of hybrid scene generation}
\label{sec:hybrid_scene_gen}
To address \textbf{Q2} (\textit{Can our hybrid scene generation method help the policy generalize to unseen and challenging scenes?}), We evaluate and compare the zero-shot scene generalization ability of the traversal policies trained with different scene datasets.

\textbf{Experiment Setting.} We collect 30 artist-designed indoor scenes that were not included in the training dataset. These scenes are categorized according to obstacle density into 15 easy and 15 hard cases. We test the policies trained with different datasets in these unseen scenes and compare their zero-shot transfer performance.

\textbf{Experiment Metrics.} We use the \textbf{Success Rate (SR, \%)} metric as mentioned in Section~\ref{sec:apf_for_humanoid}.

\textbf{Baselines.}
We choose the following methods as baselines:
\begin{itemize}
\item \textbf{Base:} We use 3D-FRONT scenes only.
\item \textbf{Base + Syn-Partial-Easy:} \textbf{Base} combined with a subset of moderate-difficulty ($\leq 0.6$) synthesized scenes.
\item \textbf{Base + Syn-Full-Easy:} \textbf{Base} combined with the full set of moderate-difficulty ($\leq 0.6$) synthesized scenes.
\item \textbf{Base + Syn-Full-Hard:} \textbf{Base} combined with the full set of high-difficulty ($0.4 \sim 1.0$) synthesized scenes.
\end{itemize}

\textbf{Experiment Results}
The results shown in Table~\ref{tab:succ_easy_hard} confirm that procedural obstacle generation is crucial to enhance generalization, as it systematically scales both the volume and the difficulty of training data. Despite significant performance improvement from the expansion of obstacle diversity, saturation in gains can be observed once common layout patterns are sufficiently learned. The key breakthrough emerges from training in scenes with novel complexities and tighter constraints, which induce the policy to develop the superior capabilities necessary for complex terrains.

\begin{table}[!t]
\centering
\resizebox{0.48 \textwidth}{!}{
    \begin{tabular}{
            >{\centering\arraybackslash}p{2.5cm} 
            >{\centering\arraybackslash}p{1.0cm}
            >{\centering\arraybackslash}p{1.0cm}
            >{\centering\arraybackslash}p{1.0cm}
            >{\centering\arraybackslash}p{1.0cm}
        }%
        \toprule
           \textbf{Real-world} & \textbf{Crouch-}  & \textbf{Hurdle-}  & \textbf{Side-} & \textbf{Crouch-}   \\
            \textbf{Exteroception} & \textbf{Pass}  & \textbf{Pass}  & \textbf{Pass} & \textbf{Hurdle}   \\
        \midrule
       Voxel Grids & 1/5  & 3/5  & 2/5 & 2/5  \\
       Elevation Maps & 3/5  & 3/5  & 1/5 & 2/5  \\
        \midrule
       HumanoidPF (Ours)     & \textbf{4/5}  & \textbf{5/5}  & \textbf{5/5} & \textbf{4/5 }   \\
        \bottomrule
    \end{tabular}
    }
    \caption{\textbf{Validation of HumanoidPF for real world transfer.} Success rate on four challenging real-world scenes.
    }
\vspace{-1.8em}
\label{tabel: real_world_ext}
\end{table}

\subsection{Validation of HumanoidPF for real world transfer}
\label{sec:sim2real_transfer} 
To address \textbf{Q3} (\textit{Can the HumanoidPF for observation $OBS_{Field}$ help the sim-to-real transfer?}), we compare the traversal capability of our method with other perceptual representations under real-world conditions.

\textbf{Experiment Setting.} We construct a set of representative cluttered indoor traversal scenarios in the real world, covering crouching under low-hanging obstacles, hurdling over ground objects, and negotiating narrow-side passages.
These scenarios are deliberately designed to stress perceptual reliability by evaluating whether the learned policy could remain collision-free traversal under real-world conditions.

\textbf{Experiment Metrics.} We use the \textbf{Success Rate (SR)} metric as mentioned in Section~\ref{sec:apf_for_humanoid}.

\textbf{Baselines.} We distill the same specialists to two different perceptual representation as baselines:
\begin{itemize}
\item \textbf{Voxel Grids:} Distilling to voxel-based visuomotor policy.
\item \textbf{Multi-layer Elevation Maps:} Distilling to two-layer-elevation visuomotor policy.
\end{itemize}

\textbf{Experiment Results.} Results are summarized in Table~\ref{tabel: real_world_ext}. Our HumanoidPF consistently outperforms voxel-based approaches across diverse environments. For example, on hurdle terrains, the voxel-based method often exhibits instability because it is sensitive to fine-grained geometric details of obstacles and to noise in task-irrelevant regions. In contrast, HumanoidPF remains largely unaffected by such disturbances and shows behavior that aligns more closely with simulation, further confirming its robustness in sim-to-real transfer.
Qualitative results of our method are shown in Figure~\ref{fig:teaser} and Figure~\ref{fig:more_visualization}.

\section{Conclusion}
\label{sec:conclusion}
In this work, we address collision-free humanoid traversal in cluttered indoor scenes. We introduce HumanoidPF, an informative representation to encode humanoid–obstacle relationships for RL-based traversal skill learning, advancing toward real-world domestic humanoid robot application. Despite these advances, several limitations remain. Our current framework does not yet exploit contact-rich interactions, such as leaning on obstacles or stepping onto support surfaces, and generalization to entirely unseen, highly unstructured, and extremely cluttered scenes remains an open challenge.

\bibliographystyle{IEEEtran}
\bibliography{IEEEabrv}
\end{document}